\documentclass[11pt,a4paper]{article}
\usepackage{naaclhlt2018}
\usepackage{times}
\usepackage{latexsym}
\usepackage{amsmath}
\usepackage{booktabs}
\usepackage{multirow}
\usepackage{array}
\usepackage{xcolor}
\usepackage{url}
\usepackage{graphicx}

\newcolumntype{C}[1]{>{\centering\let\newline\\\arraybackslash\hspace{0pt}}m{#1}}

\aclfinalcopy % Uncomment this line for the final submission
\def\aclpaperid{1085} %  Enter the acl Paper ID here

\title{Simple and Effective Semi-Supervised Question Answering}

\author{Bhuwan Dhingra\thanks{\enskip Equal Contribution} \qquad Danish Pruthi\footnotemark[1] \qquad Dheeraj Rajagopal\footnotemark[1] \\
School of Computer Science \\
Carnegie Mellon University, Pittsburgh, USA \\
\texttt{\{bdhingra, ddanish, dheeraj\}@cs.cmu.edu}
}

\date{}

\begin{document}
\maketitle
\begin{abstract}
Recent success of deep learning models for the task of extractive Question Answering (QA) is hinged on the availability of large annotated corpora. However, large domain specific annotated corpora are limited and expensive to construct. In this work, we envision a system where the end user specifies a set of base documents and only a few labelled examples. Our system exploits the document structure to create cloze-style questions from these base documents; pre-trains a powerful neural network on the cloze style questions; and further fine-tunes the model on the labeled examples. We evaluate our proposed system across three diverse datasets from different domains, and find it to be highly effective with very little labeled data. We attain more than 50\% F1 score on SQuAD and TriviaQA with less than a thousand labelled examples. We are also releasing a set of 3.2M cloze-style questions for practitioners to use while building QA systems\footnote{\url{http://bit.ly/semi-supervised-qa}}.%achieve more than 50\% F1 score using merely 1\% training data, and 

%This simple process is highly effective, achieving significant gains across three diverse datasets. \dancomment{Rephrase the last line + add key results in one line}

\end{abstract}
%\dancomment{add abstract}

\section{Introduction}

Deep learning systems have shown a lot of promise for extractive Question Answering (QA), with performance comparable to humans when large scale data is available. However, practitioners looking to build QA systems for specific applications may not have the resources to collect tens of thousands of questions on corpora of their choice. At the same time, state-of-the-art machine reading systems do not lend well to low-resource QA settings where the number of labeled question-answer pairs are limited (c.f. Table~\ref{tab:SQuAD}).
%Hence, low-resource QA, where the number of labeled question answer pairs are limited, is important but state-of-the-art machine reading systems perform poorly in such a setting (c.f. Table~\ref{tab:SQuAD}). 
Semi-supervised QA methods like \citep{yang2017semi} aim to improve this performance by leveraging unlabeled data which is easier to collect.

In this work, we present a semi-supervised QA system which requires the end user to specify a set of base documents and only a small set of question-answer pairs over a subset of these documents. Our proposed system consists of three stages. First, we construct cloze-style questions (predicting missing spans of text) from the unlabeled corpus; next, we use the generated clozes to pre-train a powerful neural network model for extractive QA \citep{clark2017simple,dhingra2016gated}; and finally, we fine-tune the model on the small set of provided QA pairs.

Our cloze construction process builds on a typical writing phenomenon and document structure: an introduction precedes and summarizes the main body of the article. Many large corpora follow such a structure, including Wikipedia, academic papers, and news articles. We hypothesize that we can benefit from the un-annotated corpora to better answer various questions -- at least ones that are lexically similar to the content in base documents and directly require factual information.
%\dancomment{write our hypothesis about the kind of information that is missing}

We apply the proposed system on three datasets from different domains -- SQuAD \citep{rajpurkar2016SQuAD}, TriviaQA-Web \citep{JoshiTriviaQA2017} and the BioASQ challenge \citep{tsatsaronis2015overview}. We observe significant improvements in a low-resource setting across all three datasets. For SQuAD and TriviaQA, we attain an F1 score of more than 50\% by merely using 1\% of the training data. Our system outperforms the approaches for semi-supervised QA presented in \citet{yang2017semi}, and a baseline which uses the same unlabeled data but with a language modeling objective for pretraining. In the BioASQ challenge, we outperform the best performing system from previous year's challenge, improving over a baseline which does transfer learning from the SQuAD dataset.
% the system by ~8\% for factoid questions and ~2\% for list questions. 
Our analysis reveals that questions which ask for factual information and match to specific parts of the context documents benefit the most from pretraining on automatically constructed clozes.
%on SQuAD, we report an F1 score of $50.4\%$ using under $1\%$ of the training data ($454$ questions). This method also outperforms the approaches for semi-supervised QA presented in \citet{yang2017semi}, and a baseline which uses the same unlabeled data but with a language modeling objective for pretraining. For TriviaQA (web) corpus, we see significant gains across various splits when compared against the baseline model. Further, we achieve an F1 score of 0.55 by merely using 746 labelled questions. 

\section{Related Work}

% There are three common paradigms to deal with the low-resource setting in supervised machine learning.

\textbf{Semi-supervised learning} augments the labeled dataset $L$ with a potentially larger unlabeled dataset $U$.
% , usually by adding a loss term which estimates $p(x)$ in addition to $p(y|x)$. In the context of QA, however, it is unclear why modeling the likelihood of the passage would be beneficial to the task of answer extraction. Instead, 
\citet{yang2017semi} presented a model, GDAN, which trained an auxiliary neural network to generate questions from passages by reinforcement learning, and augment the labeled dataset with the generated questions to train the QA model. Here we use a much simpler heuristic to generate the auxiliary questions, which also turns out to be more effective as we show superior performance compared to GDAN. Several approaches have been suggested for generating natural questions \citep{tang2017question,subramanian2017neural,song2017unified}, however none of them show a significant improvement of using the generated questions in a semi-supervised setting. Recent papers also use unlabeled data for QA by training large language models and extracting contextual word vectors from them to input to the QA model \citep{salant2017contextualized,peters2018deep,McCann2017LearnedIT}. The applicability of this method in the low-resource setting is unclear as the extra inputs increase the number of parameters in the QA model, however, our pretraining can be easily applied to these models as well.
% Our work also derives motivation from \citet{bajgar2016embracing}, who showed that automatically generating a large but noisy collection of questions leads to significant performance gains on the Children's Book Test (CBT) dataset \citep{hill2015goldilocks}.

\textbf{Domain adaptation} (and \textbf{Transfer learning}) leverage existing large scale datasets from a source domain (or task) to improve performance on a target domain (or task). For deep learning and QA, a common approach is to pretrain on the source dataset and then fine-tune on the target dataset \citep{chung2017supervised,golub2017two}. \citet{DBLP:conf/bionlp/WieseWN17} used SQuAD as a source for the target BioASQ dataset, and \citet{kadlec2016finding} used Book Test \citep{bajgar2016embracing} as source for the target SQuAD dataset. \citet{mihaylov2017neural} transfer learned model layers from the tasks of sequence labeling, text classification and relation classification to show small improvements on SQuAD. All these works use manually curated source datatset, which in themselves are expensive to collect. Instead, we show that it is possible to automatically construct the source dataset from the same domain as the target, which turns out to be more beneficial in terms of performance as well (c.f. Section \ref{sec:experiments}). 
% \textbf{Transfer learning} is similar to domain adaptation, except knowledge is transferred between skills (such as POS tagging and reading comprehension) rather than between domains. \citet{mihaylov2017neural} transfer learned model layers from the tasks of sequence labeling, text classification and relation classification to show small improvements on SQuAD.
Several cloze datasets have been proposed in the literature which use heuristics for construction \citep{hermann2015teaching,onishi2016did,hill2015goldilocks}.
% The CNN / DailyMail datasets \citep{hermann2015teaching} were created using a similar heuristic to our process for cloze construction. 
We further see the usability of such a dataset in a semi-supervised setting.
% Other automatically generated cloze datasets have also been proposed \citep{hill2015goldilocks,onishi2016did}.

\section{Methodology}

Our system comprises of following three steps:

\textbf{Cloze generation:} Most of the documents typically follow a template, they begin with an introduction that provides an overview and a brief summary for what is to follow. We assume such a structure while constructing our cloze style questions. When there is no clear demarcation, we treat the first $K\%$ (hyperparameter, in our case 20\%) of the document as the introduction. While noisy, this heuristic generates a large number of clozes given any corpus, which we found to be beneficial for semi-supervised learning despite the noise.
%For each sentence in the summary, we retrieve the most relevant passage from the rest of the document based of tf-idf \dancomment {Modify}. From the summary-sentence pair, we create triples of $(P, S, a)$, passage $P$, sentence from the summary $S$ and the potential answer phrase $a$. 

We use a standard NLP pipeline based on Stanford CoreNLP\footnote{https://stanfordnlp.github.io/CoreNLP/} (for SQuAD, TrivaQA and PubMed) and the BANNER Named Entity Recognizer\footnote{http://banner.sourceforge.net} (only for PubMed articles) to identify entities and phrases. %We then select answer candidates if the phrase is one of noun phrase, verb phrase, adjective phrase or a named entity. Next we match the candidate answers from introduction to the candidate answers in the individual paragraphs. We consider it a \emph{match}, when the answer candidate in the introduction appears in the paragraph (from the rest of the document). 
Assume that a document comprises of introduction sentences $\{q_1, q_2, ... q_n\}$, and the remaining passages $\{p_1, p_2, .. p_m\}$. Additionally, let's say that each sentence $q_i$ in introduction is composed of words $\{w_1, w_2, ... w_{l_{q_i}}\}$, where $l_{q_i}$ is the length of $q_i$. We consider a $\text{match} (q_i, p_j)$, if there is an exact string match of a sequence of words $\{w_k, w_{k+1}, .. w_{l_{q_i}}\}$ between the sentence $q_i$ and passage $p_j$. If this sequence is either a noun phrase, verb phrase, adjective phrase or a named entity in $p_j$, as recognized by CoreNLP or BANNER, we select it as an answer span $A$. Additionally, we use $p_j$ as the passage $P$ and form a cloze question $Q$ from the answer bearing sentence $q_i$ by replacing $A$ with a placeholder. As a result, we obtain passage-question-answer ($P, Q , A$) triples (Table~\ref{tab:example_cloze} shows an example). As a post-processing step, we prune out $(P, Q, A)$ triples where the word overlap between the question (Q) and passage (P) is less than 2 words (after excluding the stop words).
%Jaccard overlap score between words in the question and the answer bearing sentence in the passage is below a threshold \dancomment{Add the exact value}.

\begin{table}[!ht]
\small
\centering
\begin{tabular}{| p{7cm} |} 
\hline
\emph{Passage (P) :}  Autism is a neurodevelopmental disorder characterized by impaired \textbf{social interaction}, verbal and non-verbal communication, and ...%restricted and repetitive behavior. Parents usually notice .... typically before age three.
%notice signs in the first two years of their child's life. These signs often develop gradually, though some children with autism reach their developmental milestones at a normal pace and then regress. %The diagnostic criteria require that symptoms become apparent in early childhood, typically before age three.
\end{tabular}

\begin{tabular}{| p{7cm} |} 
\hline
\emph{Question (Q) }: People with autism tend to be a little aloof with little to no $\rule{1cm}{0.15mm}$.
\end{tabular}

\begin{tabular}{| p{7cm} |} 
\hline
\emph{Answer (A) }: social interaction \\
\hline
\end{tabular}
\vspace{-2mm}
\caption{An example constructed cloze.}
\label{tab:example_cloze}
\vspace{-2mm}
\end{table}

The process relies on the fact that answer candidates from the introduction are likely to be discussed in detail in the remainder of the article. In effect, the cloze question from the introduction and the matching paragraph in the body forms a question and context passage pair. We create two cloze datasets, one each from Wikipedia corpus (for SQuAD and TriviaQA) and  PUBMed academic papers (for the BioASQ challenge), consisting of 2.2M and 1M clozes respectively. From analyzing the cloze data manually, we were able to answer 76\% times for the Wikipedia set and 80\% times for the PUBMed set using the information in the passage. In most cases the cloze paraphrased the information in the passage, which we hypothesized to be a useful signal for the downstream QA task.

We also investigate the utility of forming subsets of the large cloze corpus, where we select the top passage-question-answer triples, based on the different criteria, like i) jaccard similarity of answer bearing sentence in introduction and the passage ii) the tf-idf scores of answer candidates and iii) the length of answer candidates. However, we empirically find that we were better off using the entire set rather than these subsets.

\textbf{Pre-training:} We make use of the generated cloze dataset to pre-train an expressive neural network designed for the task of reading comprehension. We work with two publicly available neural network models -- the GA Reader \citep{dhingra2016gated} (to enable comparison with prior work) and BiDAF + Self-Attention (SA) model from \citet{clark2017simple} (which is among the best performing models on SQuAD and TriviaQA). After pretraining, the performance of BiDAF+SA on a dev set of the (Wikipedia) cloze questions is 0.58 F1 score and 0.55 Exact Match (EM) score. This implies that the cloze corpus is neither too easy, nor too difficult to answer.

\textbf{Fine Tuning:} We fine tune the pre-trained model, from the previous step, over a small set of labelled question-answer pairs. As we shall later see, this step is crucial, and it only requires a handful of labelled questions to achieve a significant proportion of the performance typically attained by training on tens of thousands of questions.

\section{Experiments \& Results}
\label{sec:experiments}

\subsection{Datasets}

We apply our system to three datasets from different domains. \textbf{SQuAD} \citep{rajpurkar2016SQuAD} consists of questions whose answers are free form spans of text from passages in Wikipedia articles. We follow the same setting as in \citep{yang2017semi}, and split $10\%$ of training questions as the test set, and report performance when training on subsets of the remaining data ranging from $1\%$ to $90\%$ of the full set. We also report the performance on the dev set when trained on the full training set ($1^\ast$ in Table \ref{tab:SQuAD}). 
% We apply our semi-supervised system to two existing neural network models -- the GA Reader \citep{dhingra2016gated} for comparison to the methods of \citet{yang2017semi}, and BiDAF + Self-Attention (SA) model from \citet{clark2017simple} which is the highest performing model on SQuAD with publicly available code\footnote{\url{https://github.com/allenai/document-qa}}. 
We use the same hyperparameter settings as in prior work. We compare and study four different settings: 1) the Supervised Learning (\textbf{SL}) setting, which is only trained on the supervised data, 2) the best performing \textbf{GDAN} model from \citet{yang2017semi}, 3) pretraining on a Language Modeling (\textbf{LM}) objective and fine-tuning on the supervised data, and 4) pretraining on the \textbf{Cloze} dataset and fine-tuning on the supervised data. The LM and Cloze methods use exactly the same data for pretraining, but differ in the loss functions used. We report F1 and EM scores on our test set using the official evaluation scripts provided by the authors of the dataset.

\textbf{TriviaQA}~\cite{JoshiTriviaQA2017} comprises of over 95K web question-answer-evidence triples. Like SQuAD, the answers are spans of text. Similar to the setting in SQuAD, we create multiple smaller subsets
% (comprising of 1\%, 5\%, 10\%, 20\%, 50\% and 90\%)
of the entire set. For our semi-supervised QA system, we use the BiDAF+SA model~\cite{clark2017simple} -- the highest performing publicly available system for TrivaQA. Here again, we compare the supervised learning (\textbf{SL}) settings against the pretraining on \textbf{Cloze} set and fine tuning on the supervised set. We report F1 and EM scores on the dev set\footnote{We use a sample of dev questions, which is the default setting for the code by~\citet{clark2017simple}. Since our goal is only to compare the models, this is not problematic.}.

We also test on the \textbf{BioASQ 5b dataset}, which consists of question-answer pairs from PubMed abstracts. We use the publicly available 
system\footnote{\url{https://github.com/georgwiese/biomedical-qa}} from \citet{DBLP:conf/bionlp/WieseWN17}, and follow the exact same setup as theirs, focusing only on factoid and list questions. For this setting, there are only $899$ questions for training. Since this is already a low-resource problem we only report results using 5-fold cross-validation on all the available data. We report Mean Reciprocal Rank (MRR) on the factoid questions, and F1 score for the list questions. %The reader is referred to \citep{DBLP:conf/bionlp/WieseWN17} for details.

% \subsection{BioASQ}
%  These QA pairs can be classified into one of factoid, list, yes-no or summary questions. For our experiments, we focus mainly on the factoid and the list questions, based on our baseline extractive QA system \citet{DBLP:conf/bionlp/WieseWN17}. Out of the total 1799 questions, there are 413 factoid and 486 list questions clearly qualifying as a low-resource setting. For pre-training, we used the open-access subset of PubMed corpus which consists of about 300,000 full-text articles of biomedical scholarly documents \footnote{https://www.ncbi.nlm.nih.gov/pmc/tools/openftlist/}. For our experiments, we follow the exact procedure mentioned in \citet{DBLP:conf/bionlp/WieseWN17}, with the same hyperparameters. The pretraimed model that we used for Our experiments show that pretraining improves the previous baseline substantially. The results are shown in table \ref{tab:res_bioasq}. 

\begin{table*}[!htbp]
\setlength\tabcolsep{1.5pt}
\scriptsize
\centering
\begin{tabular}{@{}C{1.4cm}C{0.9cm}C{0.71cm}C{0.71cm}C{0.71cm}C{0.71cm}C{0.71cm}C{0.71cm}C{0.71cm}C{0.71cm}C{0.71cm}C{0.71cm}C{0.71cm}C{0.71cm}C{0.71cm}C{0.71cm}C{0.71cm}C{0.71cm}@{}}
\toprule
\multirow{2}{*}{\textbf{Model}} & \multirow{2}{*}{\textbf{Method}} & \multicolumn{2}{c}{\textbf{0}}   & \multicolumn{2}{c}{\textbf{0.01}}   & \multicolumn{2}{c}{\textbf{0.05}}   & \multicolumn{2}{c}{\textbf{0.1}}    & \multicolumn{2}{c}{\textbf{0.2}}    & \multicolumn{2}{c}{\textbf{0.5}}    & \multicolumn{2}{c}{\textbf{0.9}}    & \multicolumn{2}{c}{\textbf{1}}     \\ \cmidrule(l){3-4} \cmidrule(l){5-6} \cmidrule(l){7-8} \cmidrule(l){9-10} \cmidrule(l){11-12} \cmidrule(l){13-14} \cmidrule(l){15-16} \cmidrule(l){17-18} 
                                &                                      & \textbf{F1} & \textbf{EM} &\textbf{F1} & \textbf{EM} & \textbf{F1} & \textbf{EM} & \textbf{F1} & \textbf{EM} & \textbf{F1} & \textbf{EM} & \textbf{F1} & \textbf{EM} & \textbf{F1} & \textbf{EM} & \textbf{F1} & \textbf{EM} \\ \midrule
\multicolumn{16}{c}{\textbf{SQuAD}} \\ \midrule
GA                              & SL                                   & -- & -- & 0.0882           & 0.0359           & 0.3517           & 0.2275           & 0.4116           & 0.2752           & 0.4797           & 0.3393           & 0.5705           & 0.4224           & 0.6125           & 0.4684           & --               & --               \\
GA                              & GDAN                                & -- & --  & --               & --               & --               & --               & 0.4840           & 0.3270           & 0.5394           & 0.3781           & 0.5831           & 0.4267           & 0.6102           & 0.4531           & --               & --               \\
GA                              & LM                                   & -- & -- &         0.0957         &      0.0394            &     0.3141             &    0.1856              &     0.3725             &   0.2365               &     0.4406             &    0.2983              &    0.5111              &       	0.3589           &       0.5520           &       0.3964           & --               & --               \\
GA                              & Cloze                              & -- & --  & 0.3090           & 0.1964           & 0.4688           & 0.3385           & 0.4937           & 0.3588           & 0.5575           & 0.4126           & 0.6086           & 0.4679           & 0.6302           & 0.4894           & --               & --               \\ \midrule
BiDAF+SA                        & SL                   
& -- & -- & 0.1926           & 0.1018           & 0.4764           & 0.3388           & 0.5639           & 0.4258           & 0.6484           & 0.5031           & 0.7044           & 0.5615           & 0.7287           & 0.5874           & 0.7154           & 0.8069           \\
BiDAF+SA                        & Cloze                               & \textbf{0.0682} & \textbf{0.032} & \textbf{0.5042}  & \textbf{0.3751}  & \textbf{0.6324}  & \textbf{0.4862}  & \textbf{0.6431}  & \textbf{0.4995}  & \textbf{0.6839}  & \textbf{0.5413}  & \textbf{0.7151}  & \textbf{0.5767}  & \textbf{0.7369}  & \textbf{0.6005}  & \textbf{0.7186}  & \textbf{0.8080}  \\ \midrule
\multicolumn{16}{c}{\textbf{TRIVIA-QA}} \\ \midrule
BiDAF+SA & SL & -- & -- & 0.2533 & 0.1898 & 0.4215 & 0.3566 & 0.4971 & 0.4318 & 0.5624 & 0.5077 & 0.6867 & 0.6239 & 0.7131 & 0.6617 & 0.7291 & 0.6786  \\
BiDAF+SA & Cloze & \textbf{0.1182} & \textbf{0.0729} & \textbf{0.5521} & \textbf{0.4807} & \textbf {0.6245} & \textbf {0.5614 } & \textbf{0.6506} & \textbf{0.5893} & \textbf{0.6849} & \textbf{0.6281} & \textbf{0.7196} & \textbf{0.6607}& \textbf {0.7381} &\textbf{0.6823} & \textbf{0.7461} & \textbf{0.6903}  \\ \bottomrule
\end{tabular}
    \caption{\small A holistic view of the performance of our system compared against baseline systems on SQuAD and TriviaQA. Column groups represent different fractions of the training set used for training.}
    \label{tab:SQuAD}
    \vspace{-0.15in}
\end{table*}

\subsection{Main Results}

Table \ref{tab:SQuAD} shows a comparison of the discussed settings on both SQuAD and TriviaQA. Without any fine-tuning (column $0$) the performance is low, probably because the model never saw a real question, but we see significant gains with Cloze pretraining even with very little labeled data. The BiDAF+SA model, exceeds an F1 score of $50\%$ with only $1\%$ of the training data ($454$ questions for SQuAD, and $746$ questions for TriviaQA), and approaches $90\%$ of the best performance with only $10\%$ labeled data. The gains over the SL setting, however, diminish as the size of the labeled set increases and are small when the full dataset is available. %Nevertheless, Cloze pretraining manages to improve the BiDAF+SA model to a new state-of-the-art on the TriviaQA dataset \dancomment{verify or remove this}.

\begin{table}[!htbp]
\centering
\scriptsize
\begin{tabular}{@{}lcc@{}}
\toprule
Method & Factoid MRR & List F1 \\ \midrule
SL$^\ast$ & 0.242 & 0.211 \\
SQuAD pretraining & 0.262 & 0.211 \\
Cloze pretraining & \textbf{0.328} & \textbf{0.230} \\ \bottomrule
\end{tabular}
\caption{\small 5-fold cross-validation results on BioASQ Task 5b. $^\ast$Our SL experiments showed better performance than what was reported in \citep{DBLP:conf/bionlp/WieseWN17}.} %$\dagger$Best performing system from \citep{DBLP:conf/bionlp/WieseWN17}.}
\label{tab:res_bioasq}
\vspace{-0.15in}
\end{table}

Cloze pretraining outperforms the GDAN baseline from \citet{yang2017semi} using the same SQuAD dataset splits. Additionally, we show improvements in the $90\%$ data case unlike GDAN. Our approach is also applicable in the extremely low-resource setting of $1\%$ data, which we suspect GDAN might have trouble with since it uses the labeled data to do reinforcement learning. Furthermore, we are able to use the same cloze dataset to improve performance on both SQuAD and TriviaQA datasets. When we use the same unlabeled data to pre-train with a language modeling objective, the performance is worse\footnote{Since the GA Reader uses \textit{bidirectional} RNN layers, when pretraining the LM we had to mask the inputs to the intermediate layers partially to avoid the model being exposed to the labels it is predicting. This results in a only a subset of the parameters being pretrained, which is why we believe this baseline performs poorly.}, showing the bias we introduce by constructing clozes is important.

On the BioASQ dataset (Table \ref{tab:res_bioasq}) we again see a significant improvement when pretraining with the cloze questions over the supervised baseline. The improvement is smaller than what we observe with SQuAD and TriviaQA datasets -- we believe this is because questions are generally more difficult in BioASQ. \citet{DBLP:conf/bionlp/WieseWN17} showed that pretraining on SQuAD dataset improves the downstream performance on BioASQ. Here, we show a much larger improvement by pretraining on cloze questions constructed in an \textit{unsupervised} manner from the same domain.

\begin{figure*}
	\centering
    \includegraphics[width=0.75\textwidth,trim={5mm 5mm 10mm 2mm}, clip]{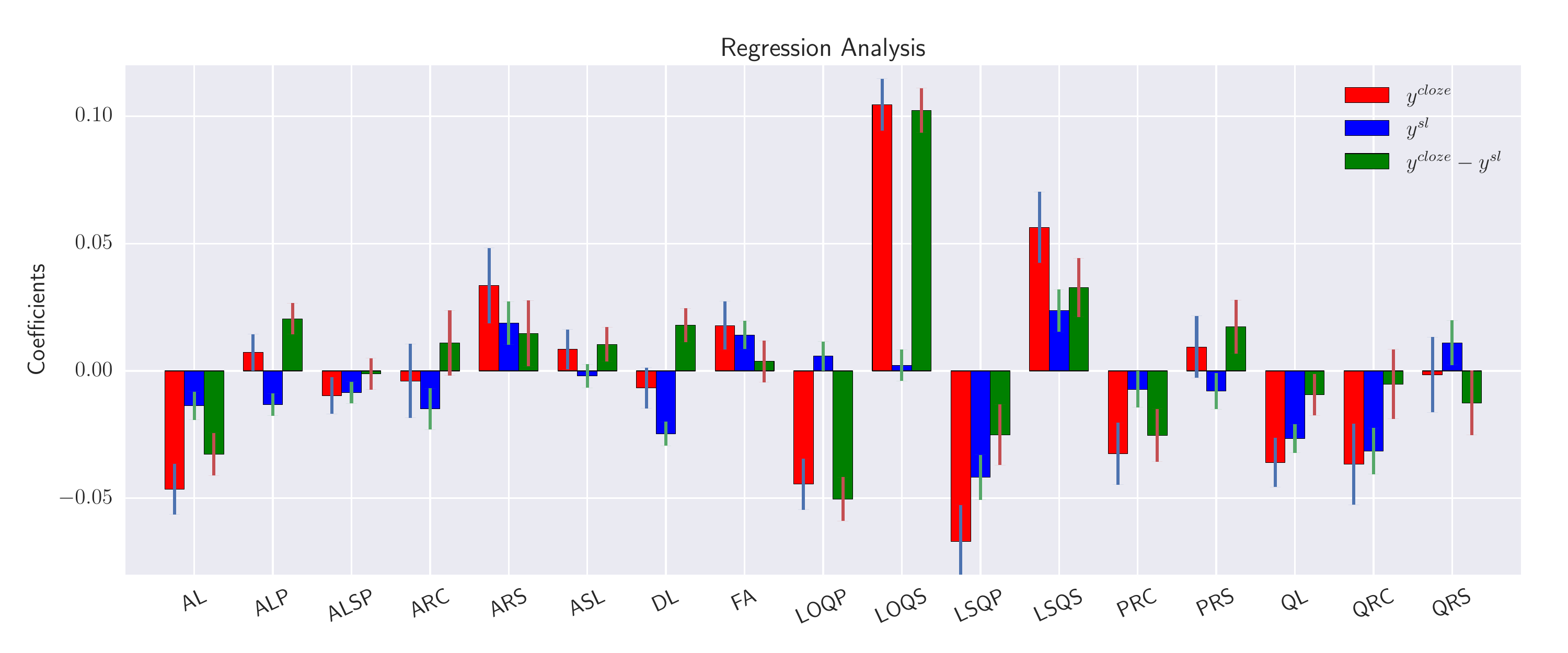}~
    \raisebox{0.095\height}{\includegraphics[width=0.25\textwidth]{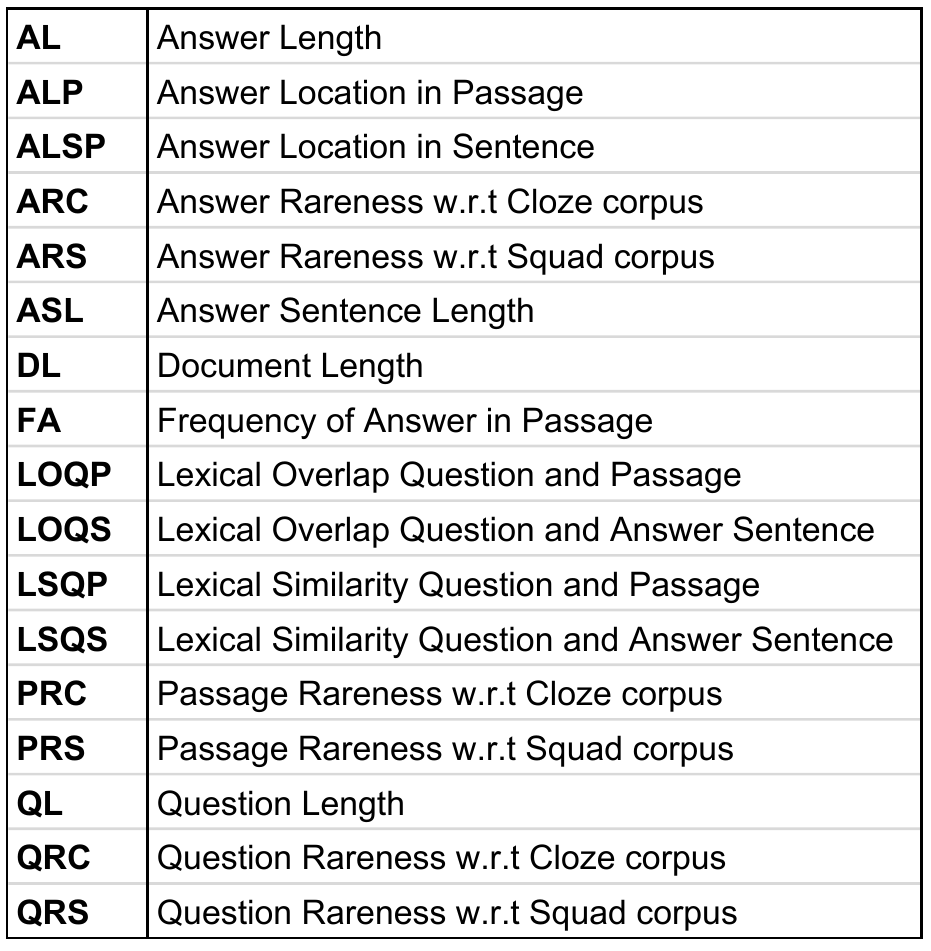}}
    \caption{\small \textbf{Left:} Regression coefficients, along with std-errors, when predicting F1 score of \textit{cloze} model, or \textit{sl} model, or the difference of the two, from features computed from SQuAD dev set questions. \textbf{Right:} Descriptions of the features.} \label{fig:regression}
\end{figure*}

% To analyze which types of questions benefit from pre-training, we automatically extract features  for each of the dev set questions in SQuAD, and fit a linear regression model to predict the F1 score for that question from these features.  

\subsection{Analysis}

\textbf{Regression Analysis:} To understand which types of questions benefit from pre-training, we pre-specified certain features (see Figure \ref{fig:regression} right) for each of the dev set questions in SQuAD, and then performed linear regression to predict the F1 score for that question from these features. We predict the F1 scores from the cloze pretrained model ($y^{\text{cloze}}$), the supervised model ($y^{\text{sl}}$), and the difference of the two ($y^{\text{cloze}}-y^{\text{sl}}$), when using $10\%$ of labeled data. The coefficients of the fitted model are shown in Figure \ref{fig:regression} (left) along with their std errors. Positive coefficients indicate that a high value of that feature is predictive of a high F1 score, and a negative coefficient indicates that a small value of that feature is predictive of a high F1 score (or a high difference of F1 scores from the two models in the case of $y^{\text{cloze}}-y^{\text{sl}}$).

% The detailed analysis is included in Appendix \ref{appendix:regression}, here we discuss the main findings. 
The two strongest effects we observe are that a high lexical overlap between the question and the sentence containing the answer is indicative of high boost with pretraining, and that a high lexical overlap between the question and the whole passage is indicative of the opposite. This is hardly surprising, since our cloze construction process is biased towards questions which have a similar phrasing to the answer sentences in context. Hence, test questions with a similar property are answered correctly after pretraining, whereas those with a high overlap with the whole passage tend to have lower performance. The pretraining also favors questions with short answers 
%(Answer Length has a negative coefficient), 
because the cloze construction process produces short answer spans. Also passages and questions which consist of tokens infrequent in the SQuAD training corpus receive a large boost after pretraining, since the unlabeled data covers a larger domain. 

\begin{figure}
    \centering
    \includegraphics[width=0.8\linewidth]{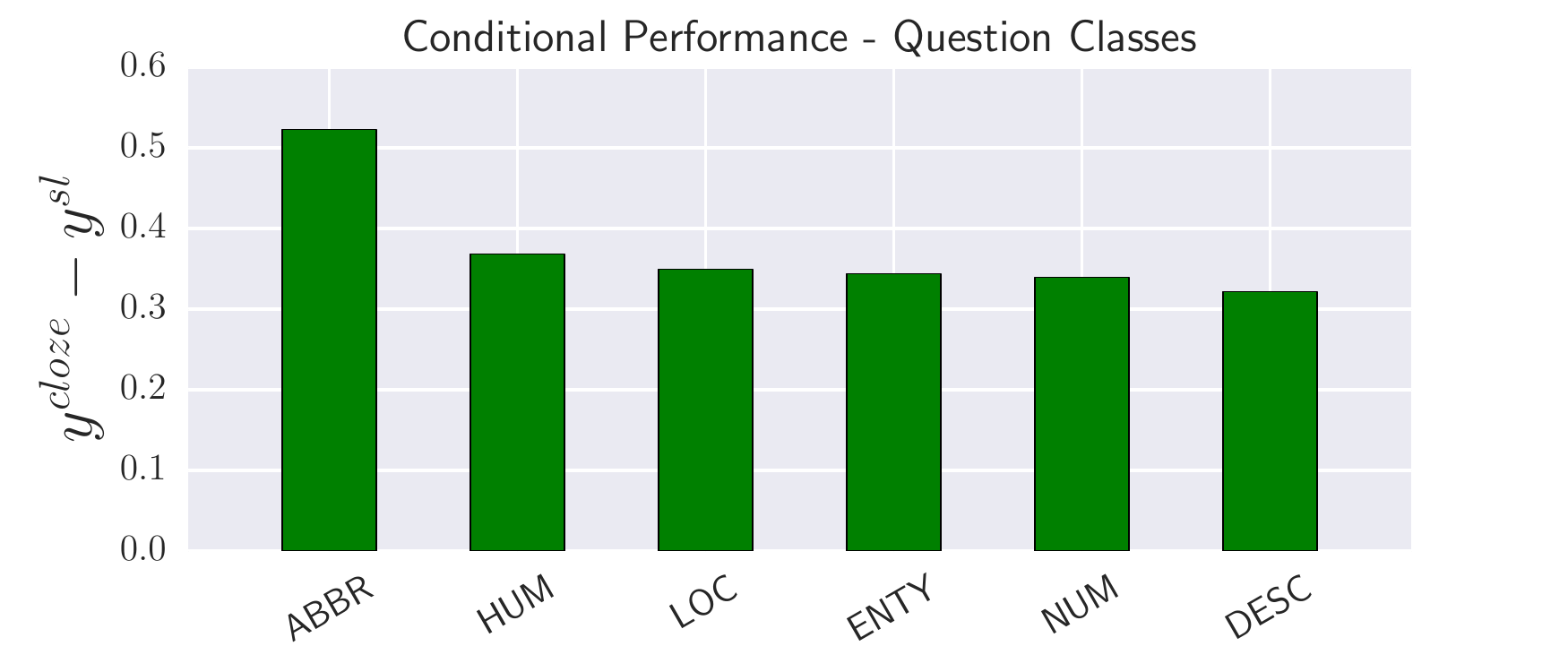}\\
%     \caption{Caption}
%     \label{fig:my_label}
% \end{figure}
% \begin{figure}
%     \centering
    \includegraphics[width=0.8\linewidth]{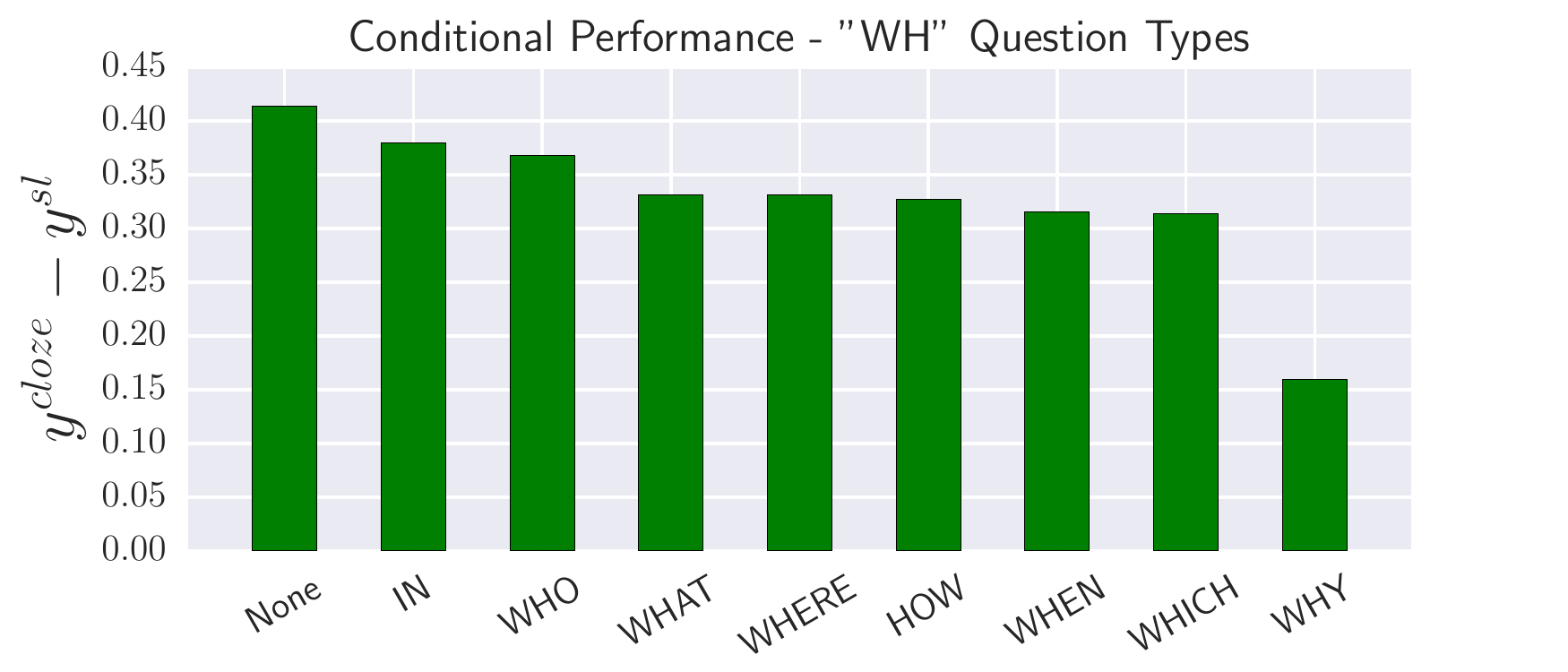}
    \caption{Performance gain with pretraining for different subsets of question types.}
    \label{fig:question_types}
\end{figure}

\textbf{Performance on question types:} Figure \ref{fig:question_types} shows the average gain in F1 score for different types of questions, when we pretrain on the clozes compared to the supervised case. This analysis is done on the $10\%$ split of the SQuAD training set. We consider two classifications of each question -- one determined on the first word (usually a wh-word) of the question (Figure \ref{fig:question_types} (bottom)) and one based on the output of a separate question type classifier\footnote{https://github.com/brmson/question-classification} adapted from \cite{li2002learning}. We use the coarse grain labels namely Abbreviation (ABBR), Entity (ENTY), Description (DESC), Human (HUM), Location (LOC), Numeric (NUM) trained on a Logistic Regression classification system . While there is an improvement across the board, we find that abbreviation questions in particular receive a large boost. Also, "why" questions show the least improvement, which is in line with our expectation, since these usually require reasoning or world knowledge which cloze questions rarely require.

% We also compute the average gain in F1 for categories of questions (described in Appendix \ref{appendix:qtypes}). We find that largest gains are for questions which ask for abbreviations, and questions which start with ``in", ``who", and ``what". The smallest gain is for questions which start with ``why".

\section{Conclusion}
In this paper, we show that pre-training QA models with automatically constructed cloze questions improves the performance of the models significantly, especially when there are few labeled examples.  
The performance of the model trained only on the cloze questions is poor, validating the need for fine-tuning.
% We evaluate this across three datasets - SQuAD, TriviaQA and BioASQ, and show gains in the low-resource setting for all three. %We also perform regression analysis to study what question types benefit from cloze style training, and 
Through regression analysis, we find that pretraining helps with questions which ask for factual information located in a specific part of the context. 
For future work, we plan to explore the active learning setup for this task -- specifically, which passages and / or types of questions can we select to annotate, such that there is a maximum performance gain from fine-tuning. We also want to explore how to adapt cloze style pre-training to NLP tasks other than QA. 

\section*{Acknowledgments}
Bhuwan Dhingra is supported by NSF under grants CCF-1414030 and IIS-1250956 and by grants from Google. Danish Pruthi and Dheeraj Rajagopal are supported by the DARPA Big Mechanism program under ARO contract W911NF-14-1-0436.

\bibliography{naaclhlt2018}
\bibliographystyle{acl_natbib}

\appendix

\end{document}